\documentclass{article}
\usepackage{spconf,amsmath,graphicx}
\usepackage{amsmath,amsfonts}
\usepackage{amssymb}
\usepackage{multirow}
\usepackage{subcaption}
\usepackage{cite}

\usepackage{comment}
\usepackage{color, colortbl}
\usepackage{tipa}
\usepackage{booktabs}
\usepackage{hyperref}

\title{Speaker Style-Aware Phoneme Anchoring for improved Cross-Lingual Speech Emotion Recognition}
\name{Shreya G. Upadhyay${^1}$, Carlos Busso$^{2}$, Chi-Chun Lee${^1}$}
\address{${^1}$Department of Electrical Engineering, National Tsing Hua University, Taiwan. \\ ${^2}$Language Technologies Institute, Carnegie Mellon University, USA.}

\begin{document}
%
\maketitle
\begin{abstract}

Cross-lingual \emph{speech emotion recognition} (SER) remains a challenging task due to differences in phonetic variability and speaker-specific expressive styles across languages. Effectively capturing emotion under such diverse conditions requires a framework that can align the externalization of emotions across different speakers and languages. To address this problem, we propose a speaker-style aware phoneme anchoring framework that aligns emotional expression at the phonetic and speaker levels. Our method builds emotion-specific speaker communities via graph-based clustering to capture shared speaker traits. Using these groups, we apply dual-space anchoring in speaker and phonetic spaces to enable better emotion transfer across languages. Evaluations on the MSP-Podcast (\emph{English}) and BIIC-Podcast (\emph{Taiwanese Mandarin}) corpora demonstrate improved generalization over competitive baselines and provide valuable insights into the commonalities in cross-lingual emotion representation.
\end{abstract}

\section{Introduction}
\label{sec:intro}
\vspace{-0.1cm}

\emph{Speech emotion recognition} (SER) is essential for emotionally intelligent human-computer interaction, with applications in dialogue systems, virtual assistants, mental health monitoring, and customer service \cite{narayanan2013behavioral,acosta2009using, tawari2010speech, devillers2010real,Lotfian_2017}. While promising in controlled conditions, SER performance often drops in real-world settings due to mismatches in language, speaker traits, or recording environments. This challenge has led to growing interest in cross-domain or cross-lingual SER, which aims to build models that generalize emotion recognition across languages and domains  \cite{parthasarathy2020semi, ahn2021cross, xu2023zero}. To mitigate domain shift, approaches such as adversarial training, feature normalization, and contrastive learning have been explored \cite{Abdelwahab_2018_3, Gideon_2021, su2021conditional}. 
Anchoring-based approaches have recently gained attention as effective strategies for improving cross-domain or cross-lingual generalization in SER \cite{upadhyay2023phonetic, upadhyay2024layer, upadhyay2025mouth, upadhyay2025phonetically}. These methods align emotionally meaningful subspaces, such as phoneme-level features across corpora with different linguistic properties. By identifying linguistic units that consistently convey emotion across languages as anchors, these methods guide models to learn stable and transferable emotion representations. This targeted anchoring mitigates domain mismatch and enhances both the robustness and interpretability of emotion features, making it effective for cross-lingual SER.

While anchoring-based methods have shown promising results, they often rely on features extracted from large pretrained or self-supervised learning (SSL) models such as WavLM \cite{chen2022wavlm} and HuBERT \cite{hsu2021hubert} to represent phonetic content. Although these models provide rich and powerful embeddings, the learned representations are inherently entangled. They capture a mixture of phonetic structure, speaker identity, prosody, and even environmental noise. This complexity makes it difficult to discern which specific information is being anchored across corpora. It raises questions such as whether we are aligning phonemes, vocal style, or both. Such ambiguity limits interpretability and makes it challenging to exert fine-grained control over the transfer process, particularly in cross-lingual SER where aligning emotion-relevant cues is essential for effective generalization.

Emotional expression in speech depends not only on what is said, represented by phonemes and lexical units, but also on how it is spoken. The expressive realization of emotion is shaped by speaker-specific prosodic cues, vocal dynamics, and articulatory styles, which vary across individuals and cultures \cite{cole2015prosody, upadhyay2025mouth}. While phonemes carry important emotional information \cite{lee2004emotion, vlasenko2014modeling}, their realization can differ depending on speaker characteristics and phonological structure of the language \cite{semin2002cultural, siegert2014investigation, yu2021phonemes}. These variations are conveyed through differences in pitch, intensity, and voice quality, all of which influence how emotion is embedded in speech \cite{lee2004emotion}. Therefore, a more interpretable and controllable SER system must consider both the phoneme-level content and the speaker style with which it is delivered, especially in cross-lingual settings.

Recent trends in speech and vision have shown that disentangled representations improve interpretability and control in tasks involving complex factors such as identity, style, and content. In particular, many studies in voice conversion \cite{li2023freevc, wang2021vqmivc}, speech synthesis \cite{hsu2019disentangling}, and emotional speech editing \cite{peng2023emotalk} explicitly separate different representation spaces to enable more controllable and adaptive modeling. In this study, we adopt a dual-encoder setup based on FreeVC \cite{li2023freevc}, where one encoder models phonetic content and the other captures speaker-related characteristics. Although speaker encoders in VC models are primarily designed for speaker identification (ID), recent findings suggest that speaker embeddings trained for ID tasks still encode emotion-dependent expressivity, leading to clustering patterns that vary with emotion \cite{Ulgen_2024}. By using these encoders independently, our framework captures emotion-relevant cues from both phoneme and speaker-style dimensions. This setup allows us to investigate how shared phonemes, when articulated with similar expressive patterns can support emotion transfer across languages. We argue that effective cross-lingual emotion recognition requires aligning both what is spoken (phonetic content) and how it is expressed (speaker style).

In this work, we propose the \emph{speaker-style aware phoneme anchoring} (SAPA), a novel framework for improving cross-lingual SER. We evaluate SAPA on two diverse corpora: the MSP-Podcast (\emph{American English}) and BIIC-Podcast (\emph{Taiwanese Mandarin}) corpora. First, we analyze how emotional cues align across languages in speaker and phoneme spaces. Despite linguistic differences, speakers form emotion-specific clusters, and some phonemes appear to carry emotional cues across corpora.  These observations motivate our dual-space anchoring strategy, where the model learns from both phonetic-level acoustic patterns and speaker-style variations. By guiding learning through these complementary spaces, SAPA captures stable emotional cues that generalize across languages. In cross-lingual transfer tasks, SAPA outperforms strong baselines, achieving 59.25\% of unweighted average recall (UAR) when training with the MSP-Podcast corpus and testing with the BIIC-Podcast corpus and 56.54\% UAR for the reverse, highlighting the effectiveness of speaker-style informed phoneme anchoring.


\begin{figure*}[!ht]    
\centering
    \begin{subfigure}{.335\textwidth}
       \includegraphics[width=\textwidth]{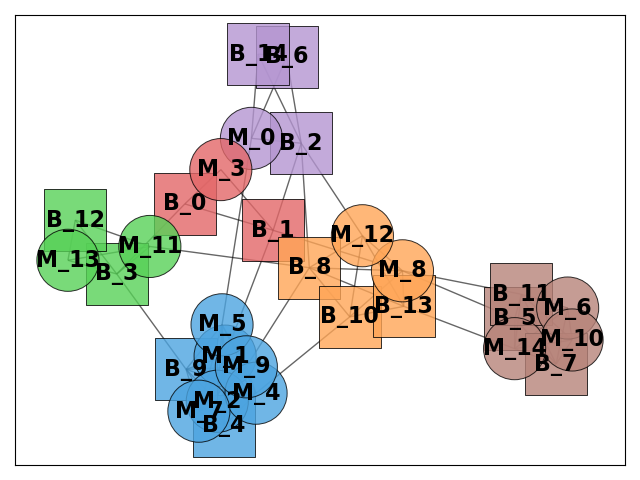}
       \subcaption{\emph{happiness}}
       \label{fig:wav-neu}
    \end{subfigure}%
    \begin{subfigure}{.335\textwidth}
        \includegraphics[width=\textwidth]{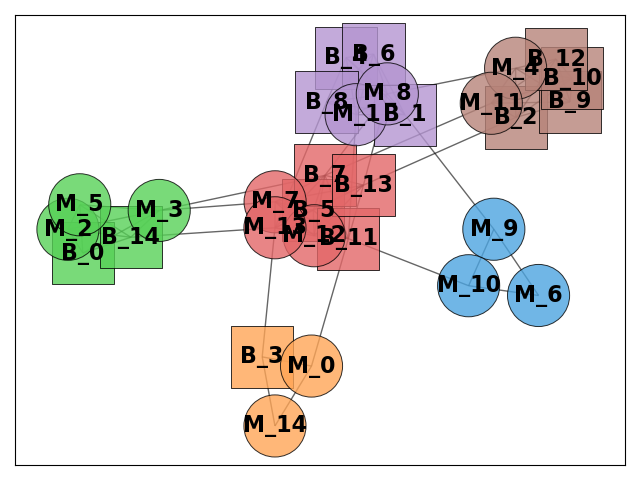}
        \subcaption{\emph{anger}}
        \label{fig:wav-hap}
    \end{subfigure}%
    \begin{subfigure}{.335\textwidth}
       \includegraphics[width=\textwidth]{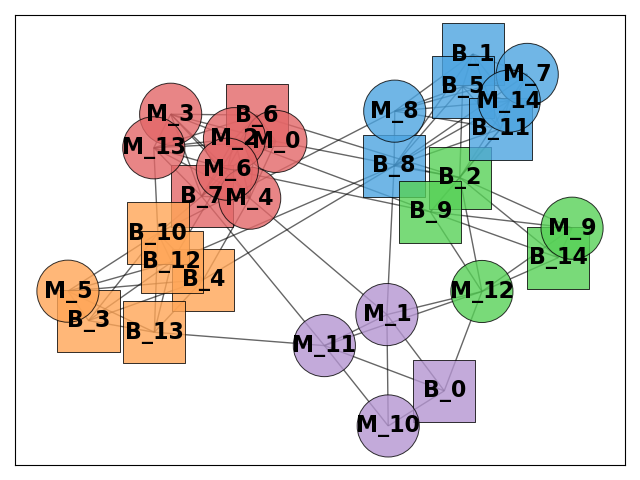}
       \subcaption{\emph{sadness}}
       \label{fig:wav-ang}
    \end{subfigure}
\caption{Speakers from the MSP-P and BIIC-P corpora are clustered using graph-based methods for each emotion category. Different colors indicate distinct speaker groups (clusters), while node shapes represent the dataset origin (circle ($\bigcirc$) for MSP-P, square ($\square$) for BIIC-P).}
\label{fig:graph_nodes}
\vspace{-0.3cm}
\end{figure*}

\section{Embedding-Level Commonality}
\label{sec:preanalyses}
\vspace{-0.1cm}

In this section, we analyze the emotion-specific commonalities across the two languages by examining both the speaker style space and the phonetic representation space.

\subsection{Affective Naturalistic Corpora}
\vspace{-0.1cm}
\textbf{The MSP-Podcast (MSP-P)} \cite{lotfian2017building} corpus comprises 324 hours of emotional \emph{American English} speech (version 1.12), collected from audio-sharing platforms. Its large scale and emotionally balanced dialogues from diverse speakers make it a valuable resource for SER research. The corpus includes annotations for primary emotions, secondary emotions, and emotional attributes.
In this study, we focus on four primary emotion categories: \textit{neutral, happiness, anger}, and \textit{sadness}, using a total of 49,018 samples with predefined training, validation, and test splits. The corpus includes phonetic information for all samples.

\smallskip
\noindent
\textbf{The BIIC-Podcast (BIIC-P)} \cite{upadhyay2023an} corpus is a SER dataset (version 1.0) in \emph{Taiwanese Mandarin}. It comprises 157 hours of podcast speech and adopts a data collection methodology similar to that of the MSP-P corpus. The dataset includes annotations for primary and secondary emotions, along with three emotional attributes. We utilize 22,799 samples focused on four primary emotional categories, using the predefined train-validation-test splits provided by the dataset creators. For phonetic information, we apply the phone alignment approach described in Upadhyay et al. \cite{upadhyay2023an}.

\subsection{Speaker-Style Encoded Commonality}
\label{sec:speaker_community}
\vspace{-0.1cm}
To capture speaker-level variations in emotional expression across corpora, we adopt a graph-based similarity clustering approach \cite{pentari2024speech, li2023speech, shirian2021compact}. This method allows us to effectively uncover communities of speakers who share similar expressive styles. We construct the graph using embeddings from the speaker encoder of the FreeVC architecture \cite{li2023freevc}, which captures speaker-specific traits such as prosody, voice quality, and articulation style. These embeddings form the basis for identifying speaker clusters that are emotion-consistent.


Let $\mathcal{S} = \{ \mathbf{z}_s^{(1)}, \mathbf{z}_s^{(2)}, \dots, \mathbf{z}_s^{(N)} \}$ be the set of speaker embeddings corresponding to $N$ speakers, extracted specifically from utterances labeled with a target emotion (e.g., \textit{anger}). We construct an undirected, weighted graph $G = (\mathcal{V}, \mathcal{E})$, where, each node $v_i \in \mathcal{V}$ corresponds to a speaker embedding $\mathbf{z}_s^{(i)}$. Each edge $e_{ij} \in \mathcal{E}$ is weighted by the cosine similarity between the speaker embeddings as shown in Eq.~\ref{eq:graph}. To focus on meaningful relationships, we retain only edges with similarity above a threshold $\tau = 0.7$.
\vspace{-0.1cm}
\begin{equation}
\vspace{-0.1cm}
    w_{ij} = \cos(\mathbf{z}_s^{(i)}, \mathbf{z}_s^{(j)}) 
    = \frac{\mathbf{z}_s^{(i)} \cdot \mathbf{z}_s^{(j)}}
           {\left\lVert \mathbf{z}_s^{(i)} \right\rVert \, \left\lVert \mathbf{z}_s^{(j)} \right\rVert}
    \label{eq:graph}
\end{equation}

We then apply Louvain clustering \cite{combe2015louvain} to identify speaker communities, where each cluster $C_k$ groups speakers with similar emotional speaking styles across languages.

To assess the quality of our emotion-specific graph-based speaker clusters, we visualize a subset of 15 speakers each from the MSP-P and BIIC-P corpora on the pre-constructed similarity graph for each major emotion. We plot the nodes using distinct shapes to represent the datasets (circles for MSP-P and squares for BIIC-P) and use colors to indicate cluster membership. As shown in Figure~\ref{fig:graph_nodes}, speakers with similar emotional expression styles tend to group together, even across languages, suggesting that the graph captures emotion-driven similarities that generalize cross-lingually.

Figure~\ref{fig:graph_nodes} shows that several clusters include speakers from both corpora, supporting our hypothesis that emotional vocal traits exhibit cross-lingual alignment. Additionally, the clustering patterns vary across emotions, some speakers group together under one emotion (e.g., \textit{sadness}) but diverge under another (e.g., \textit{anger}), indicating that the embeddings capture emotion-specific articulatory differences. These observations suggest that our graph-based approach can effectively model dynamic and interpretable speaker-style patterns across languages.


To further validate the emotion-relevance of these clusters, we compute the modularity score \( Q \) for each emotion graph, which quantifies how well the graph partitions into dense, emotion-consistent communities.

\vspace{-0.1cm}
\begin{equation}
\vspace{-0.15cm}
Q = \frac{1}{2m} \sum_{i,j} \left[ A_{ij} - \frac{k_i k_j}{2m} \right] \delta(c_i, c_j)
\end{equation}

\noindent
where \( A_{ij} \) is the edge weight between nodes \( i \) and \( j \), \( k_i \) is the degree of node \( i \), \( m \) is the total edge weight in the graph, and \( \delta(c_i, c_j) \) is 1 if nodes \( i \) and \( j \) belong to the same cluster and 0 otherwise. The modularity score \( Q \) ranges from 0 to 1, with higher values indicating better-defined and more cohesive community structures.

The modularity scores for different emotions are presented in Table~\ref{tab:modularity}. As shown, all considered emotions yield relatively high modularity values, indicating that speaker expressions are consistent and form well-defined clusters. This supports the use of emotion-specific speaker groupings to guide our dual-space phoneme anchoring and suggests that certain emotions offer more stable cues for effective cross-lingual alignment.

\begin{table}[]
\centering
\caption{Modularity scores of graph structures across different emotion categories.}
\renewcommand{\arraystretch}{1.1}
\resizebox{0.7\columnwidth}{!}{%
\begin{tabular}{c|c|c}
\toprule\specialrule{\cmidrulewidth}{0pt}{0pt}
          & Clusters & Modularity Score\\
          \hline\hline
neutral   & 5                  & 0.64       \\
happiness & 6                  & 0.69       \\
anger     & 5                  & 0.68       \\
sadness   & 5                  & 0.61      \\
\specialrule{\cmidrulewidth}{0pt}{0pt}\bottomrule    
\end{tabular}}
\label{tab:modularity}
\vspace{-0.3cm}
\end{table}

\subsection{Phoneme-Encoded Commonality}
\label{sec:phoneme_commonality}
\vspace{-0.1cm}
To assess phoneme alignment across corpora, we analyze the content encoder outputs from the FreeVC-based model \cite{li2023freevc}, focusing on phonemes that appear in both datasets under the same emotion. While earlier studies \cite{upadhyay2023phonetic, upadhyay2025phonetically} highlight the effectiveness of phoneme-level anchoring in cross-lingual SER, our objective is to verify whether these shared phonemes can serve as stable anchors in our framework, supporting emotion transfer between languages.


To analyze phoneme-level embeddings, we extract phonetic representations from the content encoder of the FreeVC-based model \cite{li2023freevc}. We perform phoneme-centered segmentation by taking a fixed 120~ms window around the midpoint of each phoneme segment, following the approach in Upadhyay et al. \cite{upadhyay2023phonetic}. These embeddings are associated with specific emotions across both corpora, for example, \( z_c^{/a/, \text{anger}} \) represents the embedding for the phoneme /a/ in the context of \textit{anger}. We identify phonemes that are shared and common across corpora within the same emotion (e.g., /\textipa{a}/ and /\textipa{i}/ under \textit{anger}) and compute cosine similarity between their embeddings to assess alignment across languages using Equation \ref{eq:sim}.


\begin{table}[]
\centering
\caption{Phonetic similarity scores (\emph{sim}) between the MSP-P and BIIC-P corpora across various emotion categories.}
\renewcommand{\arraystretch}{1.1}
\resizebox{0.9\columnwidth}{!}{%
\begin{tabular}{c|cccccc}
\toprule\specialrule{\cmidrulewidth}{0pt}{0pt}
          & /\textipa{i}/ & /\textipa{E}/ & /\textipa{@}/ & /\textipa{A},\textipa{a}/ & /\textipa{O}/ & /\textipa{u}/ \\
\hline\hline        
neutral   & 0.72                           & \textbf{0.73}                           &\textbf{0.73}                           & \textbf{0.74}                                                     & 0.70                            & 0.72                           \\
happiness & \textbf{0.74}                           & 0.71                           & 0.68                           & \textbf{0.73}                                                        & 0.69                           & 0.68                           \\
anger     & \textbf{0.72}                           & 0.71                           & 0.70                            & \textbf{0.73}                                                        & 0.71                           & \textbf{0.72}                           \\
sadness   & 0.67                           & \textbf{0.70}                           & \textbf{0.69}                           & 0.67                                                        & \textbf{0.69}                           & 0.68   \\       

\specialrule{\cmidrulewidth}{0pt}{0pt}\bottomrule    
\end{tabular}}
\label{tab:phonetic_sim}
\vspace{-0.3cm}
\end{table}

\vspace{-0.1cm}
\begin{equation}
\mathit{sim} = \cos \left( z_c^{\text{src}}, z_c^{\text{tgt}} \right)
\label{eq:sim}
\end{equation}

The similarity score $\mathit{sim}$ provides a direct measure of phoneme-embedded commonality across corpora within the same emotional context.
This similarity serves as a direct measure of phoneme-level alignment between corpora within the same emotional context. Table~\ref{tab:phonetic_sim} shows the similarity scores for shared phonemes across corpora, where higher values indicate closer phonetic realization under the same emotion. Based on these scores, Table~\ref{tab:phonetic_candidate} lists the selected phoneme anchor candidates for each emotion category. These anchors are later used as constraining factors in the phoneme-space anchoring component of our proposed method.


Overall, our analyses across the speaker and phonetic spaces reveal key insights that support cross-lingual emotion transfer. The speaker-style graph shows that speakers from different languages often cluster together under the same emotion, indicating that expressive vocal traits transcend linguistic boundaries and can be grouped based on shared emotional articulation. Complementarily, the phoneme-level analysis shows that certain phonemes consistently encode emotional information across corpora, pointing to stable phonetic patterns that generalize across languages. These insights form the foundation of our dual-space anchoring approach, demonstrating that both speaker style and phoneme content offer transferable emotional cues essential for improving generalization in cross-lingual SER.

\begin{table}[]
\centering
\caption{Emotion-wise phonetic anchor candidates derived from similarity scores in Table~\ref{tab:phonetic_sim}.}
\renewcommand{\arraystretch}{1.1}
\resizebox{0.55\columnwidth}{!}{%
\begin{tabular}{c|c}
\toprule\specialrule{\cmidrulewidth}{0pt}{0pt}
          & Vowel Phonemes                                                \\ \hline\hline
neutral   & /\textipa{E}/, /\textipa{@}/, /\textipa{A},\textipa{a}/  \\
happiness & /\textipa{i}/, /\textipa{A},\textipa{a}/                                  \\
anger     & /\textipa{A},\textipa{a}/, /\textipa{i}/, /\textipa{u}/                                   \\
sadness   & /\textipa{E}/, /\textipa{@}/, /\textipa{O}/     \\                            
\specialrule{\cmidrulewidth}{0pt}{0pt}\bottomrule    
\end{tabular}}
\label{tab:phonetic_candidate}
\vspace{-0.3cm}
\end{table}

\begin{figure*}[tbp]
  \centering
  \includegraphics[height=0.45\linewidth,width=0.87\linewidth]{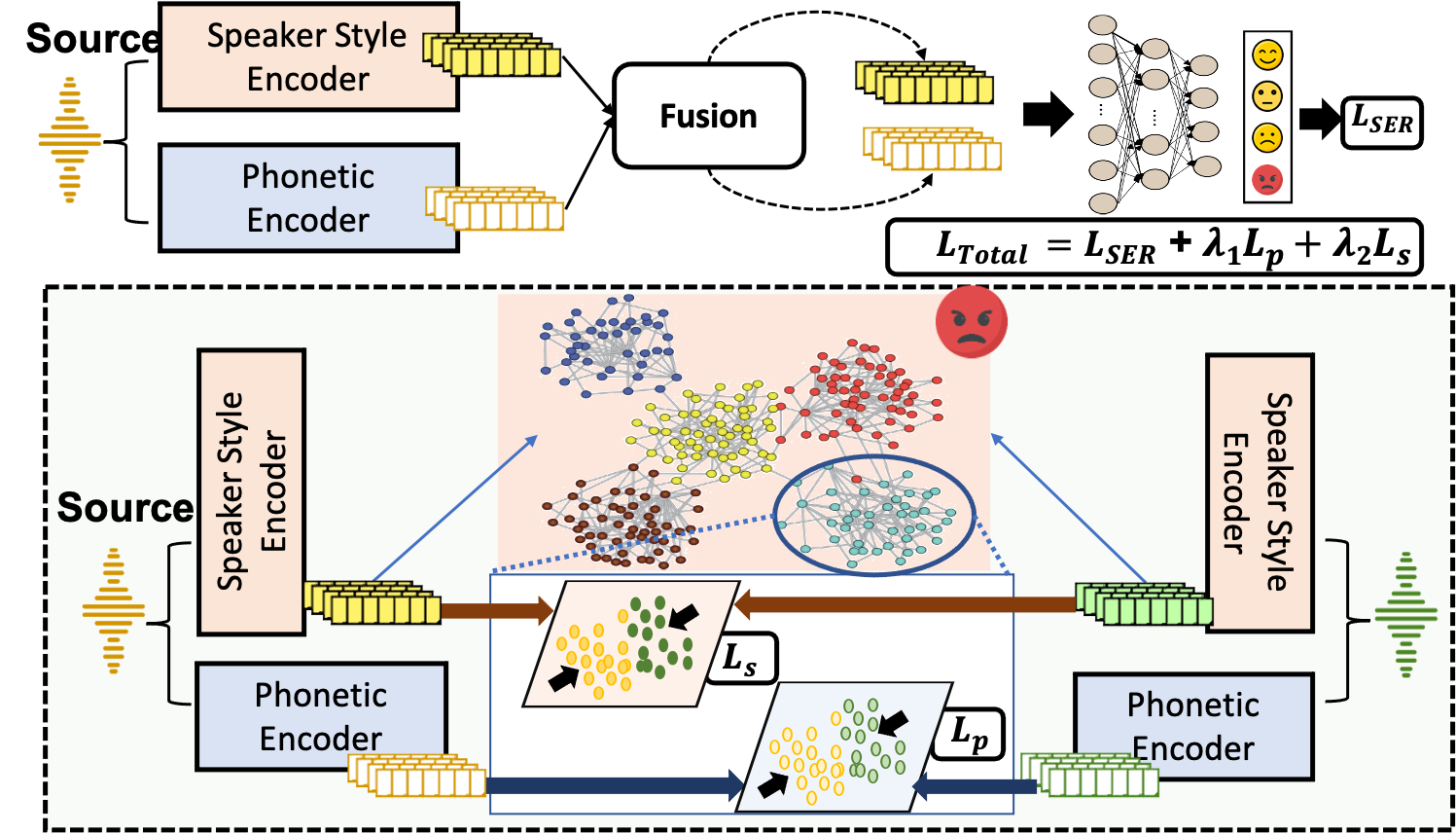}
  \caption{Proposed speaker-style aware phoneme anchoring (SAPA) architecture for cross-lingual SER.}
  \label{fig:arch}
  \vspace{-0.2cm}
\end{figure*}

\section{Speaker Style-Aware Phoneme Anchoring}
\label{sec:method}
\vspace{-0.1cm}
\subsection{Dual-Space Guided Phoneme Anchoring}
\vspace{-0.1cm}


We propose a dual-space framework that models each speech utterance using two encoders: a content encoder for phonetic structure and a speaker encoder for expressive speaker style relevant to emotion. Both are based on the FreeVC architecture \cite{li2023freevc}, producing robust phoneme- and speaker-level embeddings. This setup enables cross-corpus alignment in both spaces, guided by emotion-specific speaker communities. Phoneme-space alignment captures consistent emotional variability across corpora, while speaker-space alignment clusters speakers with similar expressive styles.

Given a speech segment \( x \), we extract two latent representations: a phonetic embedding \( \mathbf{z}_c = f_c(x) \in \mathbb{R}^{d_c} \) and a speaker embedding \( \mathbf{z}_s = f_s(x) \in \mathbb{R}^{d_s} \), where \( f_c(\cdot) \) and \( f_s(\cdot) \) denote the content and speaker encoders. The content embedding captures phoneme identity, while the speaker embedding encodes prosody, pitch, and vocal dynamics relevant to the speaker's emotions. To prepare input segments, we use the phoneme-level forced alignments and extract 240~ms windows centered around each phoneme, following the same setup used in Section \ref{sec:phoneme_commonality}. These segments are passed through both encoders to obtain fine-grained phoneme and speaker embeddings. The resulting embeddings serve as inputs to our anchoring strategy, where we enforce emotion-consistent alignment in both phonetic and speaker-style spaces.

To guide learning, we define two triplet losses—one in each latent space. The phoneme-space loss \( \mathcal{L}_p \) encourages embeddings of the same phoneme, spoken with the same emotion by speakers in similar expressive-style clusters, to remain close:
\vspace{-0.05cm}
\begin{equation}
\vspace{-0.15cm}
\mathcal{L}_{p} = \max\left(0,\,
\left\lVert \mathbf{z}_c^A - \mathbf{z}_c^P \right\rVert_2^2
- \left\lVert \mathbf{z}_c^A - \mathbf{z}_c^N \right\rVert_2^2
+ \alpha \right)
\label{eq:phoneme_loss}
\end{equation}

\noindent
where \( \mathbf{z}_c^A \) is the anchor (target phoneme embedding), \( \mathbf{z}_c^P \) is a positive sample from the same phoneme-emotion cluster, and \( \mathbf{z}_c^N \) is a negative drawn from a different emotion but the same phoneme class.

The speaker-space triplet loss \( \mathcal{L}_s \) ensures that speakers exhibiting similar expressive behavior under emotion are closely grouped in the speaker latent space:
\vspace{-0.05cm}
\begin{equation}
\vspace{-0.15cm}
\mathcal{L}_{s} = \max\left(0,\,
\left\lVert \mathbf{z}_s^A - \mathbf{z}_s^P \right\rVert_2^2
- \left\lVert \mathbf{z}_s^A - \mathbf{z}_s^N \right\rVert_2^2
+ \beta \right)
\end{equation}

\noindent
Here, \( \mathbf{z}_s^A \) and \( \mathbf{z}_s^P \) belong to speakers in the same emotion-style cluster, while \( \mathbf{z}_s^N \) is from a different cluster. Together, these losses promote cross-lingual alignment in both phoneme articulation and emotional speaking style. The margins \( \alpha \) and \( \beta \) in the triplet losses define how much farther negative samples should be from the anchor compared to positive ones. We set \( \alpha = 0.4 \) for the phoneme space and \( \beta = 0.6 \). These values are chosen based on validation performance on multi-objective loss balancing in emotion recognition.

\subsection{Emotion Classification}
\vspace{-0.1cm}
For four category emotion classification, we concatenate the pooled phonetic and speaker embeddings:

\vspace{-0.15cm}
\begin{equation}
\mathbf{z}_{\text{fused}} = \text{Concat}(\text{Pool}(\mathbf{z}_c), \text{Pool}(\mathbf{z}_s))
\end{equation}

\noindent
The fused representation, obtained by combining the phonetic and speaker embeddings, is fed into our SER model architecture. For emotion classification, we adopt a transformer-based encoder followed by four fully connected layers, consistent with the architecture used in Upadhyay et al. \cite{upadhyay2023phonetic}. This model is designed to predict one of the four emotion categories using a softmax output layer. The training objective for SER is computed using the standard cross-entropy loss.

\vspace{-0.1cm}
\begin{equation}
\mathcal{L}_{\text{SER}} = -\sum_{i=1}^{C} y_i \log(\hat{y}_i)
\end{equation}

\noindent
where $C$ is the number of emotion classes.

The final training objective integrates the SER loss with the two anchoring losses:

\vspace{-0.2cm}
\begin{equation}
\mathcal{L}_{\text{Total}} = \mathcal{L}_{\text{SER}} + \lambda_1 \mathcal{L}_{p} + \lambda_2 \mathcal{L}_{s}
\end{equation}

\noindent
where \( \lambda_1 \) and \( \lambda_2 \) are the weighting coefficients for the phoneme-space and speaker-space losses, respectively. In our experiments, we set both values to \( \lambda_1 = 0.5 \) and \( \lambda_2 = 0.5 \). Figure \ref{fig:arch} shows our proposed speaker style-aware phoneme anchoring architecture.

\section{Experimental Results}
\label{sec:exp}
\vspace{-0.1cm}
\subsection{Experimental Settings} 
\vspace{-0.1cm}
The training is optimized using the Adam optimizer in conjunction with stochastic gradient descent. A learning rate of 0.0001 and a decay factor of 0.001 are applied to support effective training. Models are trained for up to 70 epochs with a batch size of 64, and early stopping is employed to mitigate overfitting. The SER models are trained for multi-class classification across four primary emotional categories. Cross-entropy loss and triplet loss are used as the cost functions, while unweighted average recall (UAR) serves as the evaluation metric. To assess the proposed approach, the corpora are split into predefined training, validation, and testing sets.

\vspace{-0.05cm}
\subsection{Comparison with Baselines}
\vspace{-0.1cm}

As baselines for comparison, we consider three established models that align with the core ideas of our proposed approach. The first is an ensemble learning method \cite{zehra2021cross}, which aggregates predictions from multiple diverse models to improve emotion recognition performance, particularly in cross-lingual settings (denoted as Ensemble). The second is a few-shot learning approach \cite{ahn2021cross}, which enables rapid adaptation to target languages or domains using only a limited amount of labeled emotional data (Few-shot). Lastly, we include the phonetic-constraint based anchoring (PC) method \cite{upadhyay2023phonetic}, which leverages a shared phonetic representation space to guide emotion recognition across corpora.

The cross-lingual SER performance results are presented in Table \ref{tab:performance}. We denote \(<\)source corpus\(>\)→\(<\)target corpus\(>\) to indicate that the models are trained with the source corpus and tested with the target corpus. For both MSP-P→BIIC-P and BIIC-P→MSP-P tasks, the proposed SAPA-anchoring method consistently outperforms all baseline models. Specifically, SAPA achieves a UAR of 59.25\% on the MSP-P→BIIC-P transfer, representing a 1.11\% improvement over the strongest baseline, PC. Likewise, for the BIIC-P→MSP-P task, SAPA attains 56.54\% UAR, surpassing PC by 1.05\%. These improvements demonstrate that integrating speaker-style information with phoneme anchoring can enhance the generalizability of emotion recognition across languages. Compared to other baselines such as Ensemble and Few-Shot learning, SAPA shows a better performance, showing the effectiveness of our speaker-style aware phoneme anchoring approach in capturing common emotional patterns in cross-lingual SER.

\begin{table}[]
\centering
\caption{Cross-lingual SER performance in terms of UAR (\%) across baseline models and ablation studies, averaged over 10 runs. Statistical significance is reported for comparisons between the baseline models and the proposed SAPA model, denoted by asterisks (* for p \(<\) 0.1, ** for p \(<\) 0.05).}
\renewcommand{\arraystretch}{1.1}
\resizebox{0.9\columnwidth}{!}{%
\begin{tabular}{c|c|c}
\toprule\specialrule{\cmidrulewidth}{0pt}{0pt}
          & \multicolumn{2}{c}{4-Category CL-SER}   \\ \hline\hline
                   & MSP-P→BIIC-P & BIIC-P→MSP-P \\ \hline
Ensemble \cite{zehra2021cross}  & 53.90\small{**}  & 52.86 \small{**} \\
Few-Shot \cite{ahn2021cross} & 54.93\small{**} & 53.35\small{**} \\
PC \cite{upadhyay2023phonetic} & 58.14\small{*}         & 55.49\small{*}   \\  \hline   
\textbf{SAPA }     & \textbf{59.25}       &\textbf{56.54}     \\\hline
Only-S            & 51.58         & 49.93        \\
Only-P           & 56.37        & 54.86        \\\hline
SAPA-Only-S & 57.46        & 55.33        \\ 
SAPA-Only-P & 58.8 1        & 56.01        \\
\specialrule{\cmidrulewidth}{0pt}{0pt}\bottomrule    
\end{tabular}}
\label{tab:performance}
\vspace{-0.4cm}
\end{table}

\subsection{Ablation Analysis}
\vspace{-0.1cm}
To evaluate the effectiveness of our proposed SAPA method for cross-lingual SER, we conduct a series of ablation studies. In the full SAPA setup, we use a fused representation of speaker and phonetic embeddings. To isolate the contribution of each component, we first experiment with models trained using only speaker embeddings (Only-S) and only phonetic features (Only-P). Furthermore, while SAPA anchors both the speaker and phoneme spaces jointly, we explore the impact of anchoring on a single space. Specifically, under the same fusion and speaker clustering framework, we compare variants where only the speaker space is anchored (SAPA-Only-S) and where only the phoneme space is anchored (SAPA-Only-P). These ablations help disentangle the role each representation plays in cross-lingual emotion recognition performance.

The ablation results for the proposed SAPA framework are shown in Table \ref{tab:performance}. These experiments evaluate the contribution of individual components by selectively removing or isolating the speaker and phoneme spaces. Using only speaker embeddings (Only-S) results in the lowest performance, with 51.58\% for MSP-P→BIIC-P and 49.93\% for BIIC-P→MSP-P, highlighting that speaker information alone is insufficient for robust emotion recognition. In contrast, using only phoneme embeddings (Only-P) improves performance to 56.37\% and 54.86\%, indicating that phoneme structure carries more discriminative emotional cues across languages.

Further, we also analyze the impact of anchoring in individual spaces. SAPA-Only-S, which anchors only in the speaker space, achieves 57.46\% and 55.33\%, while SAPA-Only-P (anchoring only in the phoneme space) achieves 58.81\% and 56.01\% for the two tasks, respectively. These results show that anchoring in either space provides benefits for cross-lingual SER. However, the full SAPA model, which anchors both speaker and phoneme spaces, achieves the best performance with 59.25\% and 56.54\%. These findings support our central hypothesis that speaker-style aware phoneme anchoring enables more generalizable cross-lingual SER.

\vspace{-0.05cm}
\subsection{Speaker Group Transferability Analysis}
\vspace{-0.1cm}
To evaluate the effectiveness of the proposed SAPA model and specifically assess the impact of speaker-style-based grouping, we conduct a speaker group transferability analysis on the BIIC-P target corpus under the MSP-P→BIIC-P cross-corpus setting. For each emotion-specific speaker cluster graph, we aggregate all test utterances belonging to speakers within that cluster and compute the average emotion recognition accuracy over these samples.

To establish fair baselines, we compare our emotion-specific graph (w/ Emo) against two alternative grouping strategies. The first baseline uses an emotion-agnostic graph (w/o Emo), where speakers are clustered using a global graph built without considering emotion categories. The second baseline is a random grouping (Rand), where we randomly sample an equal number of utterances from speakers across the target corpus. This random sampling is repeated with multiple seeds, and the reported accuracy is averaged across runs for stability.

As shown in Table~\ref{tab:accuracy_comparison}, the emotion-specific graph-based clustering (w/ Emo) consistently outperforms both the emotion-agnostic clustering (w/o Emo) and the random baseline (Rand). Comparing w/ Emo to w/o Emo reveals that incorporating emotional context into speaker grouping leads to better emotion generalization, for example, in the case of \textit{anger}, w/ Emo achieves 72.4\% accuracy compared to 70.1\% for w/o Emo. This suggests that clustering speakers without considering emotional categories limits the model’s ability to align expressive characteristics.

When comparing w/ Emo to the Rand baseline, the benefit becomes even more pronounced. Again, using \textit{anger} as an example, w/ Emo outperforms Rand by +6.8\%, indicating that randomly grouping speakers, even with equal-sized samples, fails to capture meaningful emotional structure. This pattern persists across other expressive emotions like \textit{happiness}, where w/ Emo clusters enable stronger cross-corpus alignment. These results validate that both emotion-aware clustering and structured speaker grouping are essential for effective cross-lingual SER, particularly under speaker-style-aware phoneme anchoring.

\begin{table}[t]
\centering
\caption{SER accuracy (\%) comparison using emotion-specific (w/ Emo), emotion-agnostic (w/o Emo), and random (Rand) speaker groupings across emotion categories.}
\renewcommand{\arraystretch}{1.1}
\resizebox{0.75\columnwidth}{!}{%
\begin{tabular}{c|ccc}
\toprule\specialrule{\cmidrulewidth}{0pt}{0pt}
\textbf{Emotion} & \textbf{w/ Emo} & \textbf{w/o Emo} & \textbf{Rand}  \\ \hline\hline
neutral   & 69.7 & 68.5 & 68.1 \\
happiness & 73.3 & 69.2 & 65.1 \\
anger     & 72.4 & 70.1 & 65.6 \\
sadness   & 68.2 & 67.4 & 66.4 \\
\toprule\specialrule{\cmidrulewidth}{0pt}{0pt}
\end{tabular}}
\label{tab:accuracy_comparison}
\vspace{-0.4cm}
\end{table}

\section{Conclusion}
\vspace{-0.1cm}
\label{sec:conclusion}
This work introduces the \emph{speaker-style aware phoneme anchoring} (SAPA) framework for cross-lingual SER. Through analysis of speaker and phoneme spaces across two diverse language corpora, we find common emotion-related patterns in both speaker expressive styles and phonetic realizations. Leveraging this, SAPA employs a dual anchoring strategy to align phoneme and speaker embeddings under emotion-specific contexts. The resulting improvements in cross-lingual SER performance confirm the benefit of modeling both speaker style and phoneme-level cues. Future work will focus on expanding the emotion label set, incorporating greater speaker diversity, and exploring advanced adaptation techniques to further improve cross-lingual generalization. We also plan to validate the proposed framework on additional languages to assess its broader applicability.


\newpage

\renewcommand{\normalsize}{\fontsize{9}{10}\selectfont}
\normalsize

\bibliographystyle{IEEEbib}
\bibliography{refs}

\end{document}